\documentclass[letterpaper, 10 pt, conference]{ieeeconf}  

\IEEEoverridecommandlockouts                              

\usepackage{stfloats}
\usepackage{graphicx}
\usepackage{booktabs} 
\usepackage[T1]{fontenc}
\usepackage[utf8]{inputenc}
\usepackage{amsmath}
\usepackage{xfp}
\usepackage{amssymb}
\usepackage{pifont}
\usepackage{bbding}
\usepackage{amssymb}
\usepackage{makecell}
\usepackage{hyperref}
\usepackage[style=numeric-comp, maxnames=3, maxbibnames=10, sorting=none, giveninits=false]{biblatex} 
\usepackage{array}      
\usepackage{booktabs}   
\usepackage{multirow}
\usepackage{diagbox}  
\usepackage{xcolor}

\usepackage{arydshln}
\usepackage{authblk}
\title{\LARGE \bf
MaskVLA: Visual Masking Against Trajectory Overfitting of Vision-Language-Action Model
}

\author{
Yuxuan Jiang$^{1*}$,
Jiaying Huang$^{1,2*}$,
Ge Wang$^{1,2,3*}$,
Shenhao Yan$^{1,3}$,
Jiahao Yang$^{1,3,4}$,
Chengsi Yao$^{1,3,5}$,\\
Qi Liu$^{1,3}$,
Qing Zhao$^{1,3}$,
Shuguang Cui$^{2,1}$,
Yiming Zhao$^{3}$,
Yatong Han$^{3\dagger}$,
Zhen Li$^{2,1\dagger}$

\thanks{$^{*}$Equal contribution. $^{\dagger}$Corresponding authors.}
\thanks{Zhen Li: \texttt{lizhen@cuhk.edu.cn}}
\thanks{
\href{https://wge2002.github.io/MaskVLA/}{Project Page: MaskVLA}
}
}

\affil{
$^{1}$FNii-Shenzhen, The Chinese University of Hong Kong, Shenzhen, China \\
$^{2}$School of Science and Engineering, The Chinese University of Hong Kong, Shenzhen, China\\
$^{3}$Ising AI \quad
$^{4}$University of Glasgow \quad
$^{5}$School of Automation, Southeast University
}

\begin{document}

\IEEEaftertitletext{\vspace{-20pt}}
\maketitle

\thispagestyle{empty}
\pagestyle{empty}

\begin{abstract}

Vision-Language-Action (VLA) models integrate vision-language understanding with  executable robot actions, enabling end-to-end learning for robot control. However, our empirical analysis reveals that existing models exhibit severe trajectory overfitting when finetuned on limited datasets. To guide the model in effectively utilizing wrist camera information, we propose MaskVLA, a masking-based fine-tuning strategy. By randomly masking a small portion of the main camera’s visual information, the model is guided to autonomously learn more fine-grained, task-relevant, and effective visual features. This process leads to the emergence of robust policies, thereby enhancing the model’s capability to tackle complex manipulation tasks and improving its generalization performance. Our method has been comprehensively evaluated on RoboTwin 2.0, achieving an average success rate improvement of 23.2\% and 16.8\% compared to $\pi_0$ and OpenVLA-OFT, respectively. Furthermore, experiments on real-world ALOHA robots also demonstrate the effectiveness of our approach.

\end{abstract}

\section{INTRODUCTION}

VLA models have advanced rapidly, demonstrating immense potential in bridging high-level and visual information with precise robotic control. Their ability to generalize across novel scenarios highlights a transformative direction for embodied intelligence. However, VLA models suffer from severe overfitting, memorizing spurious dataset correlations rather than learning robust task semantics. 

The model's brittleness arises from two key factors. First, limited dataset scale and diversity forces reliance on extensive fine-tuning of small datasets to attain acceptable performance. Second, architectural limitations exacerbate domain shift: for instance, OpenVLA-OFT \cite{kimFineTuningVisionLanguageActionModels2025a} achieves strong performance, yet suffers from severe distribution shift due to mismatch between its single-view pretraining and multi-view fine-tuning, resulting in primary-camera overfitting and trajectory memorization.



We conducted an interpretability analysis using Grad-CAM \cite{Selvaraju_2019} (Fig.~\ref{fig:grad-cam}) and confirmed that OpenVLA-OFT \cite{kimFineTuningVisionLanguageActionModels2025a} suffers from a severe overfitting problem: its attention is excessively concentrated on local texture features while largely ignoring multi-view information. This finding demonstrates the model’s overreliance on superficial statistical cues rather than learning the underlying task semantics, revealing a fundamental flaw in its representation learning.
\begin{figure}[t] 
    \centering
    \includegraphics[width=\linewidth]{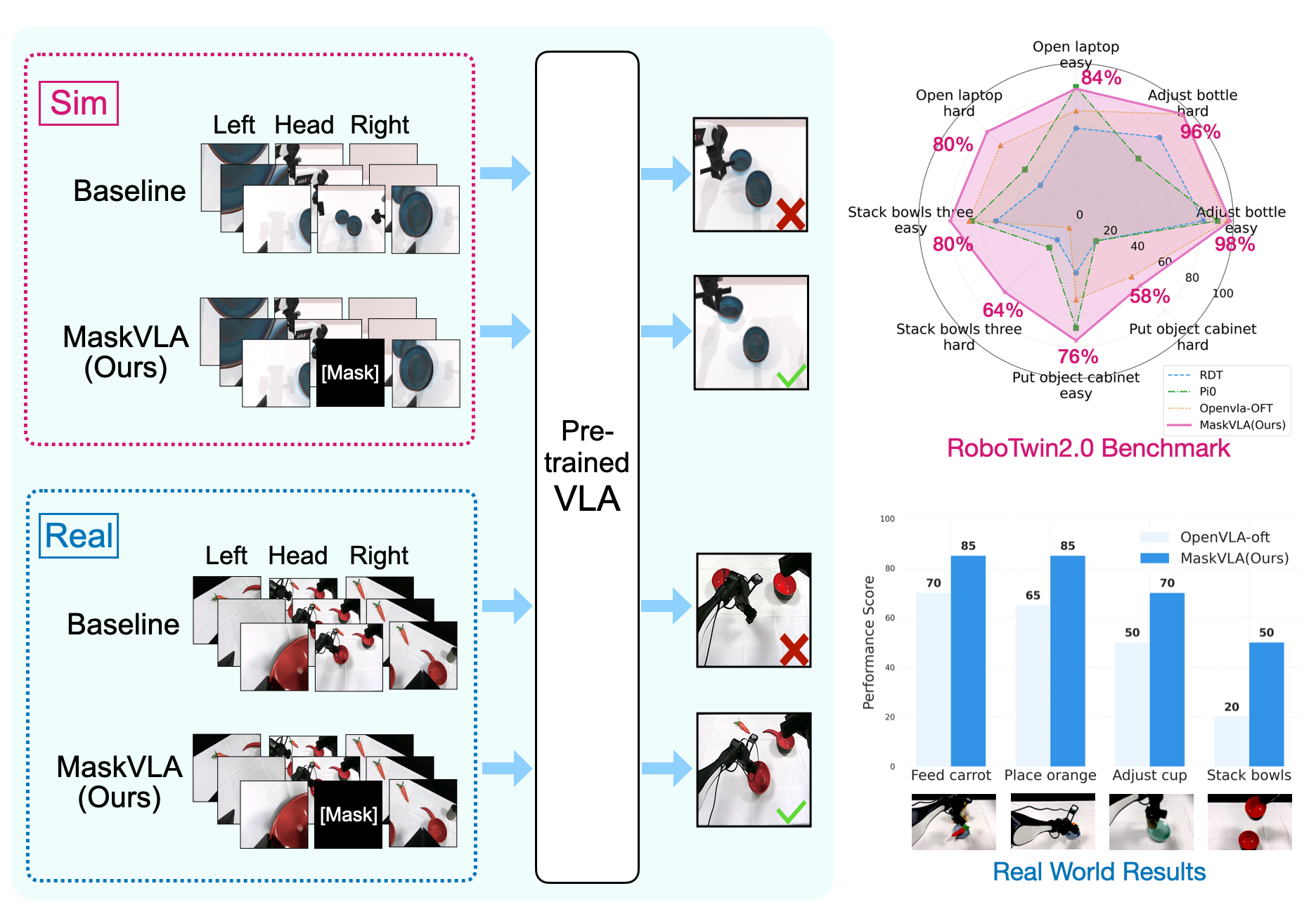} 
    \caption{
We propose \textbf{MaskVLA}, an improved fine-tuning method for multi-view VLA frameworks that significantly enhances model performance. By randomly masking the primary camera input, the method guides the model to learn robust multi-view visual features. In both RoboTwin2.0 \cite{chen2025robotwin20scalabledata}  simulation environments and real-world experiments, it demonstrates substantial improvements over OpenVLA-OFT \cite{kimFineTuningVisionLanguageActionModels2025a}  across multiple manipulation tasks.
    }
    \label{fig:intro}
\end{figure}
Although most VLA training pipelines already incorporate standard visual data augmentation techniques (e.g., random cropping, rotation, blur, and color jitter; see Table~\ref{tab:baseline_aug}), these augmentations alone are insufficient to resolve the problem. In practice, VLA models still struggle to build a stable global understanding of the scene. Instead, they often overfit to demonstration trajectories, relying on memorized spatial patterns rather than reasoning about object relationships and task goals. As a result, policies frequently exhibit brittle behaviors during execution, such as grasping at incorrect locations (e.g., closing the gripper near but not on the target object), failing to re-localize objects after small perturbations, and lacking the ability to recover from intermediate failures.

Multi-view VLAs require coordinated use of main and wrist cameras for complex tasks. While the main camera provides global context, the wrist camera becomes essential when handling close-range occlusions. Current methods, however, fail to dynamically integrate these views. Instead of predefined switching rules, the model should autonomously learn when and how to use each camera.

Inspired by the effective masking mechanisms in VLMs and LLMs, we propose MaskVLA, a novel fine-tuning method designed for multi-view VLA models. During fine-tuning, it randomly masks a small portion of the primary camera input, while during inference, it takes the complete multi-view images as input. This approach encourages the model to focus more on wrist camera information, thereby learning more view-balanced and robust visual features.

Extensive experiments in both simulation and real-world environments validate the efficacy of our method (Fig.~\ref{fig:intro}). In RoboTwin2.0 benchmark \cite{chen2025robotwin20scalabledata}, our approach significantly outperforms state-of-the-art methods such as OpenVLA-OFT \cite{kimFineTuningVisionLanguageActionModels2025a} and $\pi_0$ \cite{black$p_0$VisionLanguageActionFlow2024a}. For real-world validation, we deploy our method on an ALOHA dual-arm robot system \cite{zhao2023learning}, and the experimental results consistently demonstrate its superior effectiveness and practical applicability.

In summary, we make three major contributions:
\begin{itemize}
\item We identify the critical issue of overfitting in SOTA VLA fine-tuning methods, using Grad-CAM to visually confirm the phenomenon. 
\item We propose MaskVLA, a novel fine-tuning framework to leverage multi-view visual cues, enabling more robust representation learning and precise action generation.
\item Experiments in both simulation and real-world confirm the method's effectiveness in matigating the problem.
\end{itemize}

\begin{figure*}[t] 
    \centering
    \includegraphics[width=\textwidth]{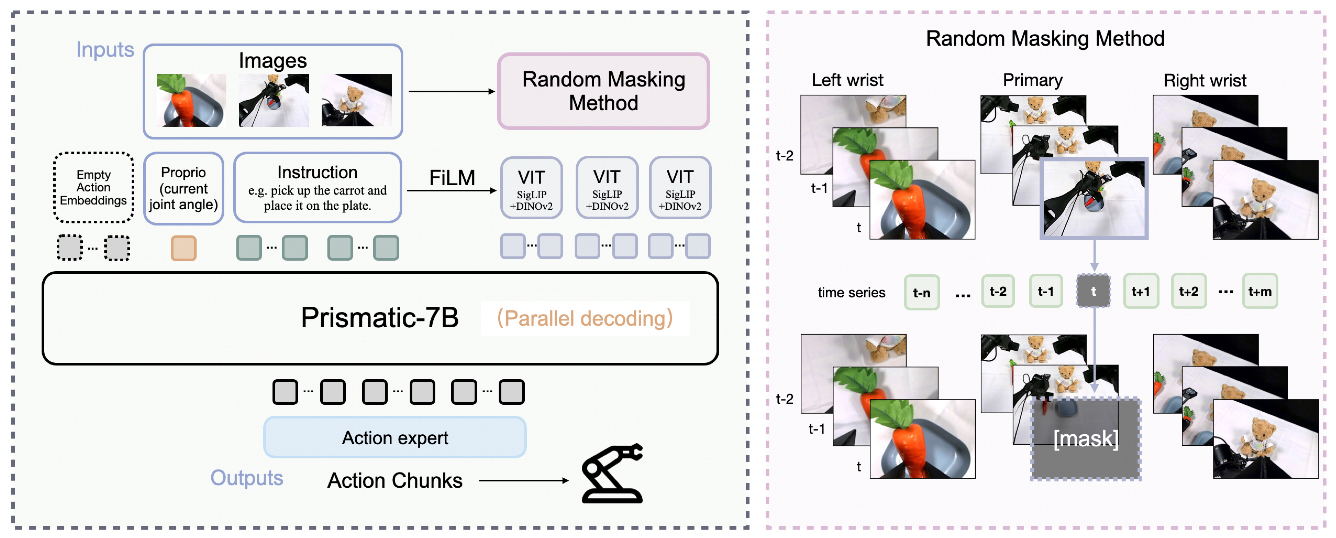}
    \caption{
  \textbf{Architecture of our MaskVLA.} During the fine-tuning stage, a small portion of the input from the primary camera (e.g., 10\%) is masked out. During inference, the model receives complete multi-camera inputs. This masking strategy reduces the model's dependency on the primary camera and guides more robust and generalized feature learning.
    } 
    \label{fig:overview}
\end{figure*}

\section{RELATED WORKS}
\subsection{Visuomotor Policy Learning}

The foundation of end-to-end robot control is built upon visuomotor policy learning methods, which map high-dimensional visual observations to low-level robot actions. A wide range of architectures have been proposed, including ACT \cite{zhao2023learning}, which employs a Transformer to predict future action chunks; Diffusion Policy (DP) \cite{chiDiffusionPolicyVisuomotor2024}, which leverages diffusion models to generate robust action sequences; and its 3D extension DP3 \cite{ze20243ddiffusionpolicygeneralizable}, which operates directly on 3D point clouds or voxels. While highly effective in continuous control tasks, these methods typically focus on imitation learning from demonstrations without incorporating natural language instructions as a conditioning signal, which limits their ability to follow open-ended semantic commands and adapt to tasks requiring linguistic understanding.

\subsection{Vision-language-action Models}

Vision-Language-Action (VLA) models \cite{black$p_0$VisionLanguageActionFlow2024a,buUniVLALearningAct2025,kimOpenVLAOpenSourceVisionLanguageAction2024,liuRDT1BDiffusionFoundation2025,nvidiaGR00TN1Open2025,quSpatialVLAExploringSpatial2025a} bridge vision-language understanding with robotic control. Built upon large pre-trained vision-language models (VLMs), they directly map multimodal inputs—visual observations and language instructions—into executable, low-level robot actions. This enables end-to-end learning of language-conditioned visuomotor policies, promising robust generalization to novel objects, environments, and semantic instructions. Representative instances include RDT \cite{liuRDT1BDiffusionFoundation2025}, which employs vision and text encoders to extract multimodal features and utilizes a DiT architecture to model action distributions in latent space, achieving end-to-end generation from perception to action; OpenVLA \cite{kimOpenVLAOpenSourceVisionLanguageAction2024}, an open-source VLA foundation model that builds upon pre-trained vision-language models and demonstrates strong performance and generalization across a variety of real-world robotic tasks through its straightforward Transformer-based architecture; and \textbf{$\boldsymbol{\pi}_\text{0}$} \cite{black$p_0$VisionLanguageActionFlow2024a}, a foundational VLA that adopts a VLM pre-training plus Flow Matching architecture, demonstrating remarkable cross-embodiment generalization capabilities. 

However, existing works largely overlook a critical issue: VLAs fine-tuned extensively on limited datasets for tens of thousands of steps often exhibit severe trajectory memorization, a clear indicator of significant overfitting. Addressing this gap, our work not only identifies this critical overfitting pitfall but also proposes a novel approach to mitigate it, paving the way for more robust VLAs.

\subsection{VLA Fine-Tuning method}

Recent advances in VLA fine-tuning \cite{zhang2024effectivetuningstrategiesgeneralist,deyReVLARevertingVisual2025a,kimFineTuningVisionLanguageActionModels2025a,} focus on improving adaptation efficiency. Zhang et al. \cite{zhang2024effectivetuningstrategiesgeneralist} systematically study the fine-tuning strategies for Octo \cite{octomodelteam2024octoopensourcegeneralistrobot}, identifying key design choices for optimal low-data performance. ReVLA \cite{deyReVLARevertingVisual2025a} reconstructs SigLIP \cite{zhai2023sigmoidlosslanguageimage} and DINOv2 \cite{oquab2023dinov2learning} weights to improve VLA out-of-distribution robustness. OpenVLA-OFT\cite{kimFineTuningVisionLanguageActionModels2025a} achieves state-of-the-art results by efficiently transforming single-arm models into bimanual multi-view policies. However, it suffers from overfitting and poor generalization due to a mismatch between single-camera pretraining data and triple-camera fine-tuning, leading to trajectory memorization. Our approach effectively mitigates this issue.

\subsection{Masking in LLM/VLM}

Masking is a core self-supervised learning method in LLMs and VLMs. 
In text domains, BERT \cite{devlin2019bertpretrainingdeepbidirectional} significantly enhances language models' semantic understanding through bidirectional context prediction. In visual domains, MAE's \cite{he2021maskedautoencodersscalablevision} high masking ratio (e.g., 75.0\%) and block-wise masking design compel the model to learn global semantic reasoning rather than local texture replication, markedly improving the generalization capability of visual representations. Recent studies (e.g., MaskGit \cite{chang2022maskgitmaskedgenerativeimage}) have further introduced bidirectional masked prediction mechanisms into image generation, employing non-autoregressive mask-iterative reconstruction strategies to maintain generation quality while substantially increasing inference speed. These advances indicate that masked modeling is evolving into a unified paradigm bridging understanding and generation tasks.

Inspired by the approaches, we introduce this method into the fine-tuning of VLA models and propose \textbf{MaskVLA}, which enhances both the generalization capability and accuracy of the model.

\section{Methodology}


We propose an improved fine-tuning framework for multi-camera Vision-Language-Action models (VLAs). Motivated by empirical evidence of overfitting in standard fine-tuning, our core contribution is a plug-and-play Random Masking on Primary Camera View module for muti-view VLA finetune, called \textbf{MaskVLA}. It effectively regularizes the learning process, leading to significant gains on challenging downstream tasks.

\subsection{Overfitting Problem in VLA}

In generalist robot policies trained on large-scale, multi-source datasets, we observe a pronounced overfitting phenomenon. This overfitting refers to the policy's tendency to memorize spurious statistical correlations – those idiosyncratic to the composition of the training dataset (e.g., specific background textures, lighting conditions, or robot kinematic properties) – rather than learning the underlying, invariant task semantics. Consequently, despite being exposed to vast amounts of data, the model remains brittle to any distribution shift that disrupts these memorized correlations.

Our analysis reveals this overfitting originates from two sources: (1) fundamental dataset limitations including insufficient diversity and fragmentation \cite{xingShortcutLearningGeneralist2025}, which promotes learning of narrow sub-distribution features; and (2) a critical architectural deficiency in OpenVLA-OFT \cite{kimFineTuningVisionLanguageActionModels2025a} where the model—pre-trained on single-arm, single-view data—is directly fine-tuned for dual-arm, multi-view tasks without commensurate pre-training. This creates a severe domain gap, forcing the policy to default to its pre-training bias and exhibit over-reliance on the primary camera view rather than developing balanced attention across all visual inputs.

To empirically validate the observed overfitting phenomenon, we conducted an interpretability analysis of the attention distribution in the model's visual encoder. Specifically, we employed Gradient-weighted Class Activation Mapping (Grad-CAM) \cite{Selvaraju_2019} to visualize the fine-tuned OpenVLA-OFT \cite{kimFineTuningVisionLanguageActionModels2025a} model. 

\begin{figure}
    \centering
    \includegraphics[width=1\linewidth]{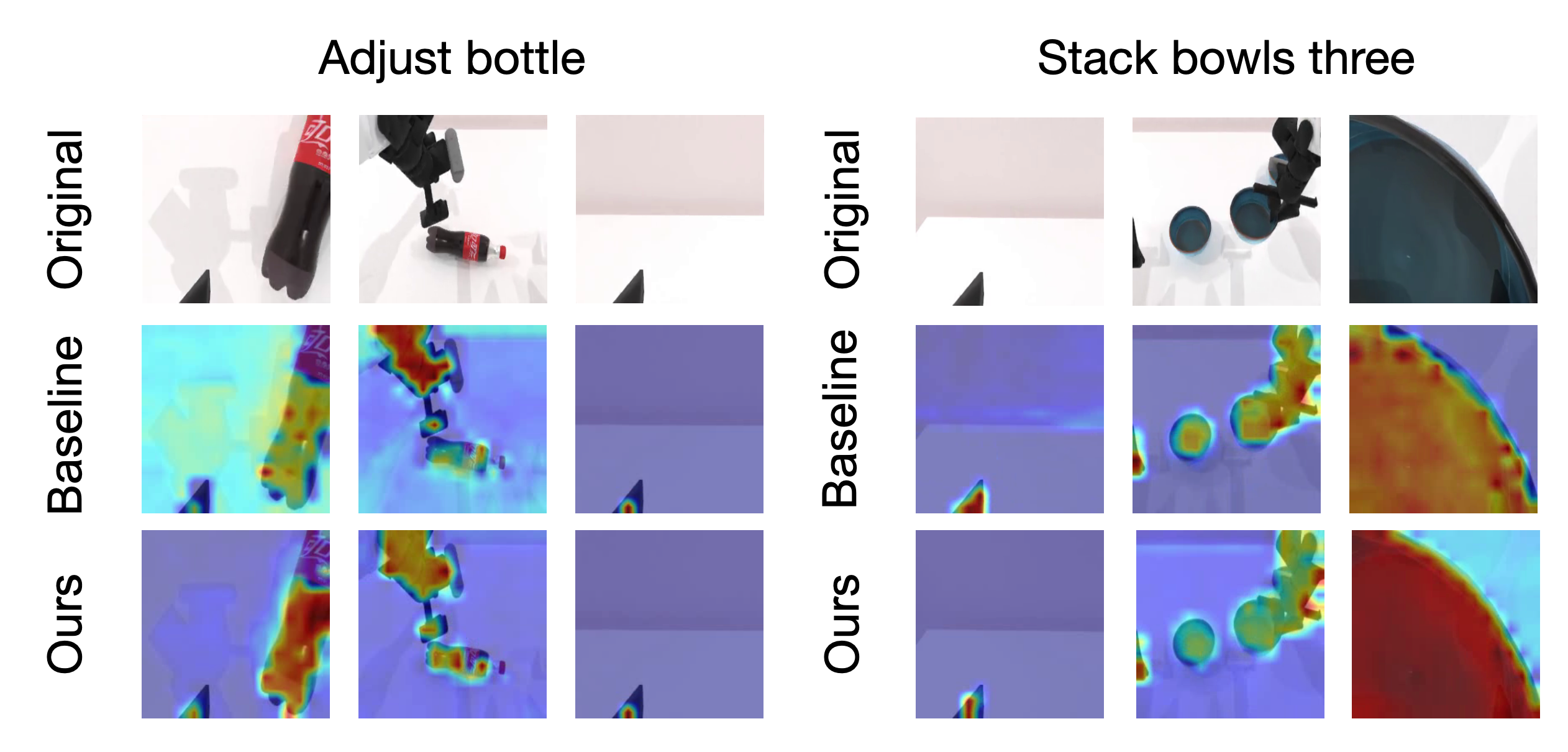}
    \caption{
    \textbf{Grad-CAM Analysis:}
 Qualitative comparisons on two RoboTwin tasks. \textbf{Row 1:} Input images. \textbf{Row 2:} Baseline (OpenVLA-OFT) shows scattered attention on irrelevant areas. \textbf{Row 3:} MaskVLA concentrates attention precisely on task-critical regions validating its superior visual grounding via masked training.}
    \label{fig:grad-cam}
\end{figure}

As shown in Fig.~\ref{fig:grad-cam}, the heatmaps clearly reveal a pronounced visual preference bias: the model's attention is highly concentrated on local texture features within the primary camera view (e.g., logos on the robot arm surface, workstation edges), while largely ignoring information from multiple auxiliary cameras. This imbalanced attention distribution confirms that the model indeed suffers from overfitting dependence on the primary view. It has learned to leverage superficial statistical features specific to this perspective rather than understanding the underlying physical principles of the task.

This finding provides computational visual evidence supporting our theoretical analysis, indicating that, under the current experimental setup, primary-view dependence primarily arises from limitations in representation learning, rather than being solely due to data volume.

To address the aforementioned over-reliance on the primary camera, we propose a novel and effective solution: Random Masking on Primary Camera View (MaskVLA). This technique is designed to explicitly discourage the policy from fixating on any single viewpoint during training. By randomly masking visual observations from the primary camera with a predetermined probability, we force the policy to learn to integrate information from all available visual streams dynamically and robustly. This approach effectively mitigates the pre-training bias and closes the domain gap by ensuring that the model does not become dependent on a potentially privileged sensory input.

\subsection{MaskVLA: Random Masking on Primary Camera View}


Our method builds upon the OpenVLA-OFT \cite{kimFineTuningVisionLanguageActionModels2025a} framework, a state-of-the-art approach for finetuning Vision-Language-Action policies. OpenVLA-OFT employs parallel decoding, continuous action prediction with an L1 regression objective, and FiLM \cite{perez2018film} modules to efficiently process multi-view visual inputs.

At each timestep $t$, the robot receives a multimodal observation defined as
\begin{equation}
O_t = \{P_t, I_p^t, I_l^t, I_r^t\},
\end{equation}
where $P_t$ represents the robot’s proprioceptive state, specifically the 14-dimensional joint angles, and $I_p^t$, $I_l^t$, $I_r^t$ correspond to RGB images from the primary (overhead) camera and the left/right wrist cameras. The primary view provides a global overview of the scene, while the wrist views offer detailed local information that is essential for precise manipulation.

Based on these observations, the VLA policy $\pi_\theta$ predicts a sequence of future actions over a horizon $H$, conditioned on the current observation and task instruction $\hat{\ell}$:
\begin{equation}\label{eq:vla_policy_seq_full}
\pi_\theta(a_{t:t+H} \mid O_t, \hat{\ell}),
\end{equation}
where $a_{t:t+H}$ denotes the predicted action sequence. This formulation allows the policy to leverage multi-view visual inputs, proprioceptive feedback, and high-level language guidance to generate coordinated manipulation behaviors.

Despite this multi-view design, existing VLA policies often develop a strong bias toward the primary camera during training, leading to poor generalization when the global view changes or becomes occluded. To mitigate this issue, we introduce \textbf{MaskVLA}, a simple stochastic masking strategy applied to the primary camera during training.

Specifically, at each discrete timestep $t_i$, the primary view is either fully masked or left intact according to a Bernoulli random variable:
\begin{equation}\label{eq:masking_bernoulli}
I_p^{t_i, \text{input}} = (1 - M_i) \odot I_p^{t_i}, \quad M_i \sim \text{Bernoulli}(\rho),
\end{equation}
where $M_i = 1$ indicates that the primary view is masked and $\rho$ denotes the masking probability. The masking operation $\mathcal{M}(\cdot)$ sets all pixel values of the primary image to zero, producing the masked image $I_p^{t_i,[\text{mask}]}$.

During training, the masked primary image $I_p^{t_i,[\text{mask}]}$ is processed together with the left- and right-wrist images $I_l^{t_i}$ and $I_r^{t_i}$ by the visual encoder to extract viewpoint-specific features $F_p$, $F_l$, and $F_r$. These features are then concatenated along the sequence-length dimension to form a fused visual representation $F$, which is fed into the VLA policy network $\pi_\theta$. The resulting observation used for policy learning is therefore
\begin{equation}
O_t = \{I_p^{t_i,[\text{mask}]}, I_l^{t_i}, I_r^{t_i}, P_t\}.
\end{equation}

By randomly masking the primary view during training, the model is encouraged to utilize wrist-view information more effectively, which is often crucial for fine-grained manipulation or situations involving occlusion. This strategy simulates scenarios where the global view is unavailable and forces the policy to learn more balanced multi-view representations. During inference, the masking operation is disabled, allowing the policy to exploit the full information from all cameras and achieve optimal performance.

\begin{table*}[t]
\vspace{5pt}
\caption{Performance Comparison of Different Methods on RoboTwin Benchmark}
\label{tab:sim-results}
\centering
\footnotesize
\begin{tabular}{
  >{\raggedright}p{2cm}  
  *{4}{                  
    >{\centering\arraybackslash}p{0.99cm}  
    >{\centering\arraybackslash}p{0.99cm}   
  }
  >{\centering\arraybackslash}p{0.9cm}
  >{\centering\arraybackslash}p{0.9cm}
  >{\centering\arraybackslash}p{0.9cm}
}
\toprule



\multirow{3}{*}{
  \setlength{\unitlength}{1pt}
  \begin{picture}(60,50)
    \put(-5,45){\line(1,-0.25){65}}  
    \put(-5,35){\makebox(0,0)[l]{\footnotesize Method}}
    \put(60,45){\makebox(0,0)[r]{\footnotesize Success Rate(\%)}}
  \end{picture}
} &


\multicolumn{2}{c}{\textbf{Adjust bottle}} & 
\multicolumn{2}{c}{\textbf{Open laptop}} & 
\multicolumn{2}{c}{\textbf{Put object cabinet}} &
\multicolumn{2}{c}{\textbf{Stack bowls three}} &
\multicolumn{3}{c}{\textbf{Average  $\mathbf{\uparrow}$}} \\
\cmidrule(lr){2-3} \cmidrule(lr){4-5} \cmidrule(lr){6-7} \cmidrule(lr){8-9} \cmidrule(lr){10-12}
 & easy & hard & easy & hard & easy & hard & easy & hard & easy & hard & all\\
\midrule
ACT \cite{zhao2023learning} & 27.0 & 14.0 & 32.0 & 0.0 & 0.0 & 0.0 & 0.0 & 0.0 & 14.8 & 3.5 & 9.1\\
DP \cite{chiDiffusionPolicyVisuomotor2024} & 97.0 & 0.0 & 49.0 & 0.0 & 42.0 & 0.0 & 63.0 & 0.0 & 62.8 & 0.0 & 31.4\\
DP3 \cite{ze20243ddiffusionpolicygeneralizable} & \textbf{99.0} & 3.0 & 82.0 & 7.0 & \underline{72.0} & 1.0 & 57.0 & 5.0 & \underline{77.5} & 4.0 & 40.8\\
RDT \cite{liuRDT1BDiffusionFoundation2025} & 81.0 & \underline{75.0} & 59.0 & 32.0 & 33.0 & 18.0 & 51.0 & 17.0 & 56.0 & 35.5 & 45.8\\
$\pi_0$ \cite{black$p_0$VisionLanguageActionFlow2024a} & 90.0 & 56.0 & \textbf{85.0} & 46.0 & 68.0 & 18.0 & 66.0 & \underline{24.0} & 77.3 & 36.0 & 56.6\\
{\fontsize{6.5}{9}\selectfont OpenVLA-OFT \cite{kimFineTuningVisionLanguageActionModels2025a} }& 96.0 & \textbf{96.0} & 70.0 & \underline{68.0} & 50.0 & \underline{50.0} & \underline{68.0} & 6.0 & 71.0 & \underline{55.0} & \underline{63.0}\\
\midrule
\textbf{MaskVLA(Ours)} & \underline{98.0} & \textbf{96.0} & \underline{84.0} & \textbf{80.0} & \textbf{76.0} & \textbf{58.0} & \textbf{80.0} & \textbf{66.0} & $\textbf{84.5}\,\textcolor{red}{\uparrow}_{\textcolor{red}{\footnotesize 7.0}}$ & $\textbf{75.0}\,\textcolor{red}{\uparrow}_{\textcolor{red}{\footnotesize 20.0}}$ & $\textbf{79.8}\,\textcolor{red}{\uparrow}_{\textcolor{red}{\footnotesize 16.8}}$\\

\bottomrule
\end{tabular}

\vspace{0.8em}

\end{table*}

\section{Simulated Experiment in RoboTwin}

Our Simulated experiments investigated the following questions:

\begin{itemize}

  \item How does the masking method improve model performance?
  \item Does the masking method help enhance the generalization ability of models?
  \item What is the optimal masking strategy?
  
\end{itemize}

\begin{figure}[t] 
    \centering
    \includegraphics[width=\linewidth]{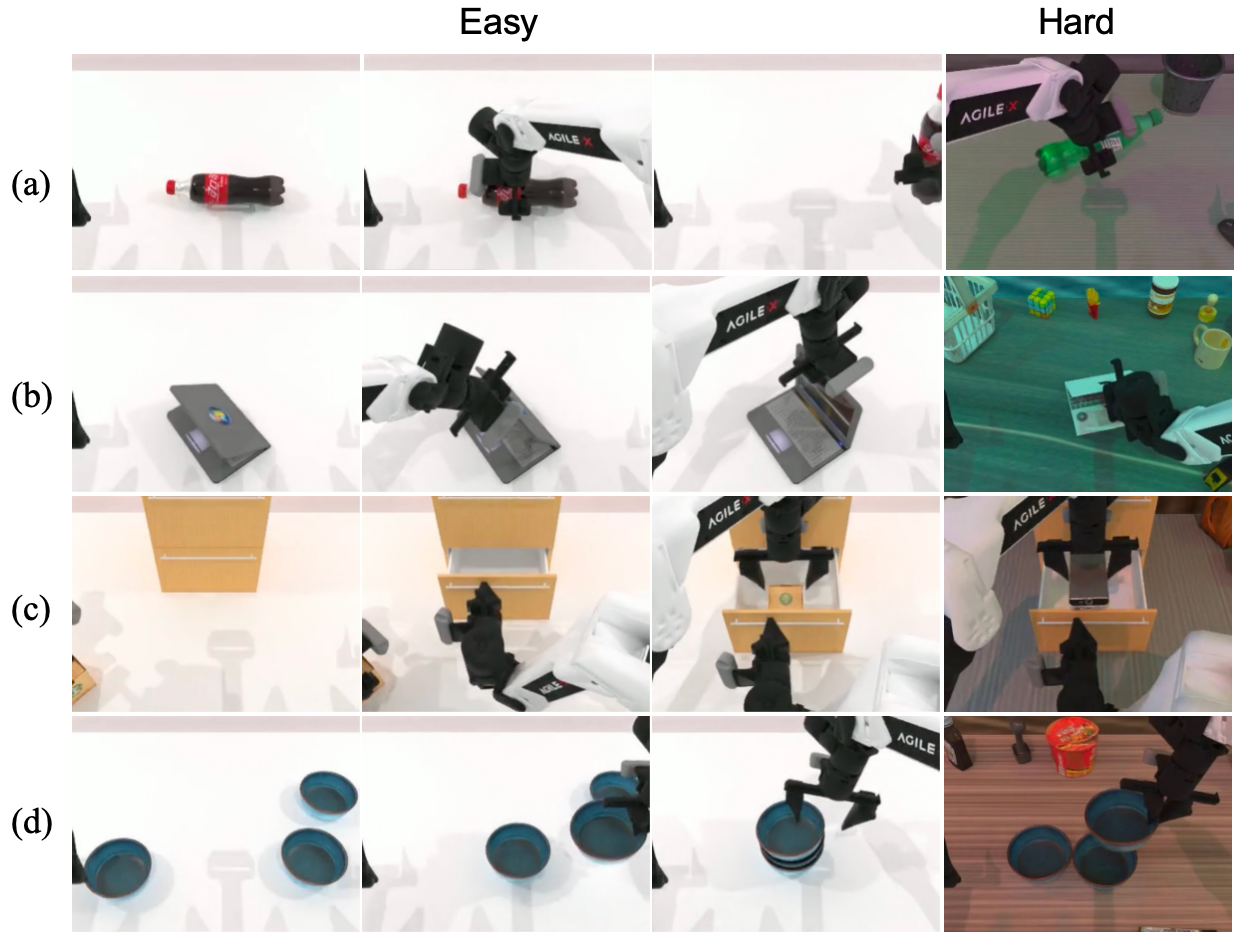}
    \caption{
  \textbf{Representative tasks from the RoboTwin2.0 benchmark.} Each row corresponds to a task: (a) Adjust bottle, (b) Open laptop, (c) Put object cabinet, and (d) Stack bowls three. The first three columns illustrate task progression in the easy setting (initial, interaction, goal), while the rightmost column shows a representative state in the hard setting with randomized object poses and layouts, evaluating policy generalization under out-of-distribution conditions.
    } 
    \label{fig:sim-experiment}
\end{figure}

\subsection{Experimental Setup}

\subsubsection{Simulation Environment}
We primarily conduct experiments on the \textbf{RoboTwin2.0} \cite{chen2025robotwin20scalabledata} simulation benchmark, using the ALOHA-AgileX dual-arm system as the manipulator to evaluate our model’s effectiveness in bimanual collaboration tasks. RoboTwin provides an excellent simulation validation environment for the proposed method. It offers a complete pipeline for data generation, algorithm verification, and evaluation. 

RoboTwin provides two environmental modes, easy and hard. Easy mode (\texttt{demo\_clean}) is in a white-background, clutter-free environment and primarily tests model task execution capability, serving as an upper bound for task execution accuracy. Hard mode (\texttt{demo\_randomized}) tests model capability by approximating the rich variability of real environments through randomly generating desktop and background textures, randomly generating lighting, serving as a more realistic evaluation environment.

\subsubsection{Task Description}
We selected four tasks require information from both the main camera and wrist cameras to complete. As shown in the pictures below Table~\ref{tab:sim-results}, the tasks are as follows:

\textbf{Adjust bottle:} requires the robot to adjust the position and orientation of a bottle. Base on the bottle's initial random orientation, the model needs to determine whether to use the left or right hand for grasping.

\textbf{Open laptop:} requires necessitating precise control of the opening angle, testing the model's ability to learn curved motion trajectories with obstructing the main camera's view.

\textbf{Put object cabinet:} involves a complex sequence of actions including opening cabinet, grasping, and placing. Different sub-steps have varying degrees of dependence on the main camera and wrist cameras respectively, testing the model's attention capability across the three cameras.

\textbf{Stack bowls three:} tests the model's generalization capability for long-horizon manipulation task. Unlike Put Object Cabinet, this task operates on identical objects.

\subsubsection{Training Protocol}

All policies are trained on the Aloha-AgileX embodiment using 50 \texttt{demo\_clean} trajectories per task. Specifically, all pre-trained policies (e.g., \textbf{$\boldsymbol{\pi}_\text{0}$} and OpenVLA-OFT) are fine-tuned on each single task based on their released open-source weights. Evaluation is conducted over 100 trials under both the \texttt{demo\_clean} (Easy) and \texttt{demo\_randomized} (Hard) settings. Unless otherwise specified, all baseline methods adopt their default data augmentation strategies as provided in the original implementations. Details of the data augmentation strategies adopted by different baseline methods are summarized in Table~\ref{tab:baseline_aug}.


\begin{table}[h]
\centering
\caption{Comparison of Data Augmentation Strategies}
\label{tab:baseline_aug}
\setlength{\tabcolsep}{4pt} 
\begin{tabular}{l ccccc}
\toprule
\textbf{Method} & \makecell{\textbf{Random}\\\textbf{Crop}\textsuperscript{1}} & \makecell{\textbf{Color}\\\textbf{Jitter}\textsuperscript{2}} & \textbf{Rotate}\textsuperscript{3} & \makecell{\textbf{Noise /}\\\textbf{Blur}\textsuperscript{4}} & \makecell{\textbf{Random}\\\textbf{Masking}\textsuperscript{5}} \\
\midrule
ACT \cite{zhao2023learning} & & & & & \\
DP \cite{chiDiffusionPolicyVisuomotor2024} & \checkmark & & & & \\
DP3 \cite{ze20243ddiffusionpolicygeneralizable} & & & & & \\
RDT \cite{liuRDT1BDiffusionFoundation2025} & & \checkmark & & \checkmark & \\ 
$\pi_0$ \cite{black$p_0$VisionLanguageActionFlow2024a} & \checkmark & \checkmark & \checkmark & & \\
{\fontsize{6.5}{9}\selectfont OpenVLA-OFT \cite{kimFineTuningVisionLanguageActionModels2025a} } & \checkmark & \checkmark & & & \\
\midrule
\textbf{Ours} & \checkmark & \checkmark & & & \checkmark \\
\bottomrule
\end{tabular}
\vspace{5pt} 
\begin{minipage}{\columnwidth}
\footnotesize 
\textsuperscript{1}Random crop and resize. 
\textsuperscript{2}Randomly alters image brightness, contrast, saturation, and hue. 
\textsuperscript{3}Random rotation within a small angle. 
\textsuperscript{4}Adds random noise or blur. 
\textsuperscript{5}Randomly masks out primary view image patches (our method).
\end{minipage}
\end{table}

\subsection{Experimental Results Analysis}
\subsubsection{Evaluation Results Analysis}
Among the selected 8 sub-tasks, the MaskVLA method achieved the best performance in 6 sub-tasks and obtained second-best performance in the remaining 2 tasks. In particular, our method achieved the highest average success rate in both Easy and Hard modes across 4 tasks, fully validating the effective improvement of the masking strategy on model performance.

In Hard mode, scene lighting and distractors vary randomly, further testing the model's scene generalization capability. Experimental results show that MaskVLA achieved an average success rate of 75.0\% in Hard mode, outperforming the baseline method OpenVLA-OFT by 20.0\%. It confirms that our method can effectively enhance the model's generalization capability.

From the specific task in Hard mode performance analysis in Table 1, our method demonstrated differentiated improvement effects across different tasks. Among them, the Stack bowls three task showed the most significant improvement, with a success rate increase of 60.0\% compared to the OpenVLA-OFT, clearly surpassing the $\pi_0$ by 42\%. For the Open laptop and Put object cabinet tasks, our method achieved performance gains of 12.0\% and 8.0\% respectively, outperforming the OpenVLA-OFT. For the Adjust bottle task, where OpenVLA-OFT already performed well, our method still maintained the same high success rate without performance degradation.

The fundamental reason why the Stack bowls three task achieved such significant improvement lies in its high requirements for position generalization capability. Although the Adjust bottle and Open laptop tasks involve random position generation, the operation targets are all single objects; in the Put object cabinet task, the variation range of object grasping positions is relatively limited, with a low degree of position generalization. In contrast, the Stack bowls three task needs to simultaneously handle three randomly distributed operation objects, placing higher demands on the model's spatial understanding ability and the generalizability of sequential operations. Therefore, our method demonstrated the most outstanding performance advantage in this most challenging position generalization scenario.

\subsection{Ablation Study}
To determine the optimal masking strategy, we designed and implemented systematic ablation experiments. All results reported are evaluated in \textbf{hard mode} to rigorously test the model's generalization to challenging and out-of-distribution scenarios. We proposed three different masking strategies for comparative evaluation:

\begin{enumerate}
\item \textbf{MaskVLA-patch:} During training, the main camera image input is segmented into patch sets after passing through the Vision Transformer, with each patch independently masked with a 10\% probability;

\item \textbf{MaskVLA-camera:} During training, at each time step, one perspective is randomly selected from the main camera view and left/right wrist camera views, and masked with a 10\% probability;

\item \textbf{MaskVLA-main-$\rho$\%:} During training, at each time step, the complete image from the main camera perspective is masked with a preset probability \textbf{$\rho$\%}.
\end{enumerate}


\begin{table*}[t]
\centering
\footnotesize
\caption{Ablation experiment results (success rates \%)}
\label{tab:ablation}
\setlength{\tabcolsep}{6pt} 
\renewcommand{\arraystretch}{1.2} 
\begin{tabular}{lcccccc}
\toprule
\multirow{2}{*}{\textbf{Method}} & \multicolumn{4}{c}{\textbf{Task}} & \multirow{2}{*}{\textbf{Avg}} \\
\cmidrule(lr){2-5}
 & Adjust bottle & Open laptop & Stack bowls & Put object & \\
\midrule
Baseline \cite{kimFineTuningVisionLanguageActionModels2025a} 
 & \textbf{96.0} & 68.0 & 6.0 & 50.0 & 55.0 \\
\midrule
MaskVLA-patch  & \textbf{96.0} & \underline{72.0} & 30.0 & 40.0 & 59.5 \\
MaskVLA-camera & \underline{94.0} & 60.0 & 58.0 & \underline{54.0} & 66.5 \\
\noalign{\vskip 2pt}
\hdashline
\noalign{\vskip 2pt}
MaskVLA-main-5\%  & \textbf{96.0} & 66.0 & \textbf{66.0} & 50.0 & 69.5 \\
MaskVLA-main-10\% & \textbf{94.0} & \textbf{80.0} & \underline{64.0} & \textbf{58.0} & \underline{74.0} \\
MaskVLA-main-20\% & 84.0 & 62.0 & 14.0 & 40.0 & 50.0 \\
\midrule
Our results & \textbf{96.0} & \textbf{80.0} & \textbf{66.0} & \textbf{58.0} & 
$\textbf{75.0}\,\textcolor{red}{\uparrow}_{\textcolor{red}{\footnotesize 20.0}}$ \\
\bottomrule
\end{tabular}

\vspace{0.5em}
\footnotesize
\textit{Note: \textbf{Bold} indicates best performance, \underline{underlined} indicates second-best. Success rates are reported for all four tasks in hard mode.}
\end{table*}

The experimental results in Table~\ref{tab:ablation} show that all models trained with masking strategies achieved average success rates approaching or exceeding the baseline method's 55.0\%, fully validating the universal effectiveness of masking methods.

Through comparative analysis, we found that the \textbf{MaskVLA-main-10\%} performed optimally, achieving task success rates above 74.0\% under appropriate mask ratio configurations, outperforming \textbf{MaskVLA-patch} and \textbf{MaskVLA-camera}. Further parameter sensitivity analysis revealed that different tasks have varying optimal mask ratios, suggesting the need for adaptive parameter tuning according to specific task characteristics.

The ablation experimental results confirmed that the \textbf{MaskVLA-main-$\rho$\%} is an efficient model optimization strategy. This method selectively occludes main perspective information during training, forcing the model to learn better integration of multi-view features, thereby significantly improving generalization performance in complex manipulation tasks.

Based on the comprehensive simulation experimental analysis above, we provide the following answers to the key questions raised at the beginning of the research:

\begin{itemize}
\item By forcing the model to integrate multi-view information during training, the masking strategy effectively addressed the overfitting problem of excessive reliance on the main camera. 

\item In tasks have high position generalization requirements, MaskVLA achieved the highest average success rate, fully demonstrating the significant enhancement effect of the masking method on model generalization capability.

\item Comparative experiments showed that our \textbf{main camera masked strategy (MaskVLA-main-$\rho$\%)} is the most effective method, selectively occludes main camera information, encouraging the model to better utilize fine manipulation information provided by wrist cameras, leading strong performance advantage in complex manipulation tasks.
\end{itemize}

\section{Experiments in Real World}

\begin{figure}[h!]
    \centering
    \includegraphics[width=\linewidth]{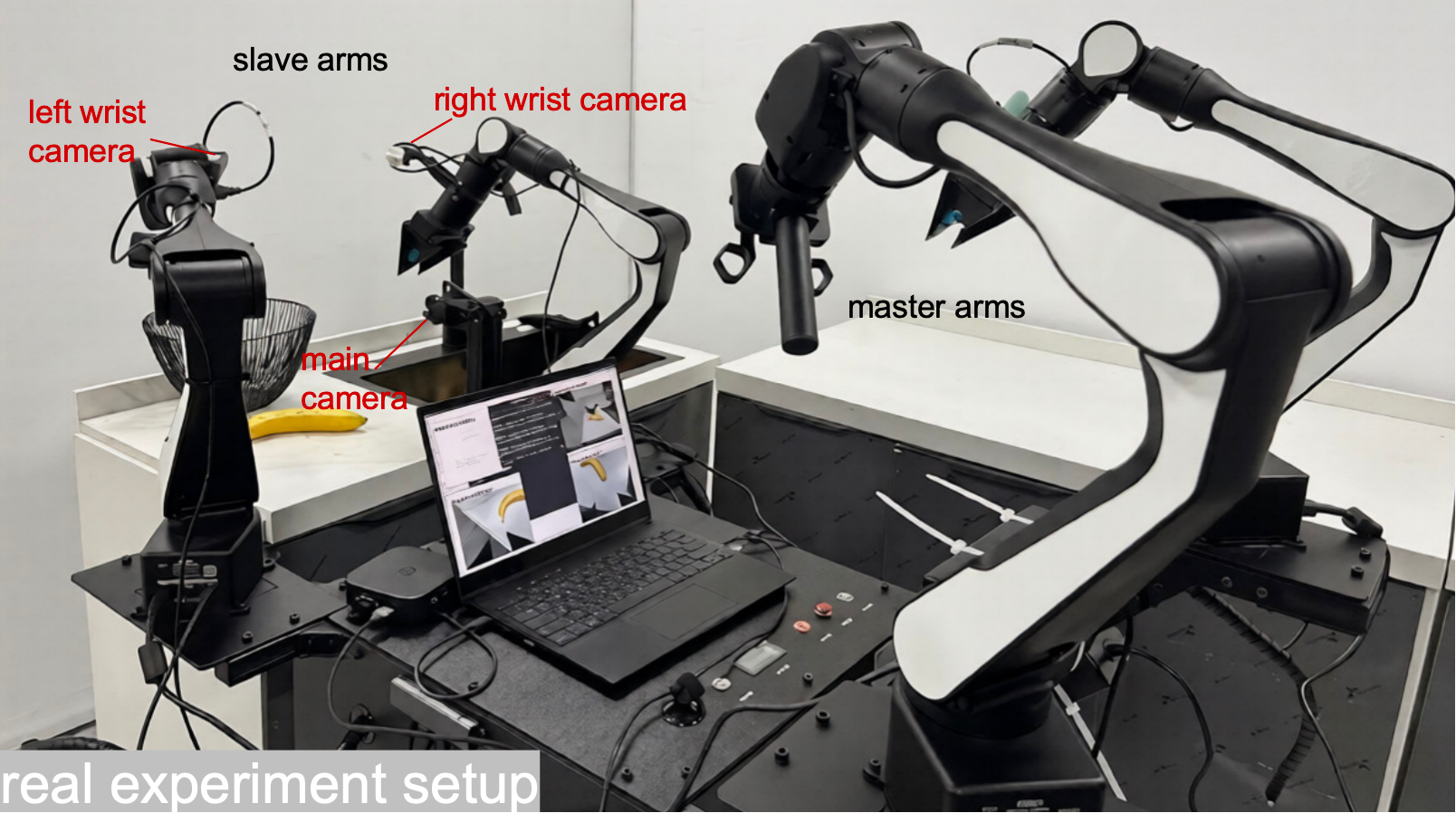} 
    \caption{
    \textbf{Real-world experimental setup.} The platform includes master and slave 6-DoF arms with 1-DoF grippers and three cameras. The master arm is used for teleoperation, while the slave arm executes tasks, enabling multi-view data collection for evaluating VLA policies.
    }
    \label{fig:real_robot_setup}
\end{figure}

\subsection{Experimental Setup}
Experiments are conducted on the Airbot Mobile ALOHA dual-arm platform (Fig.~\ref{fig:real_robot_setup}). The robot is controlled using proprioceptive and visual inputs, including 14-dimensional joint angles and images from three cameras.

We collect 50 trajectories per task. For evaluation, each model is tested on each task with 20 trials, and the success rate is used as the performance metric. Training follows the same protocol as in simulation experiments.

\begin{figure}[t] 
    \centering
    \includegraphics[width=\linewidth]{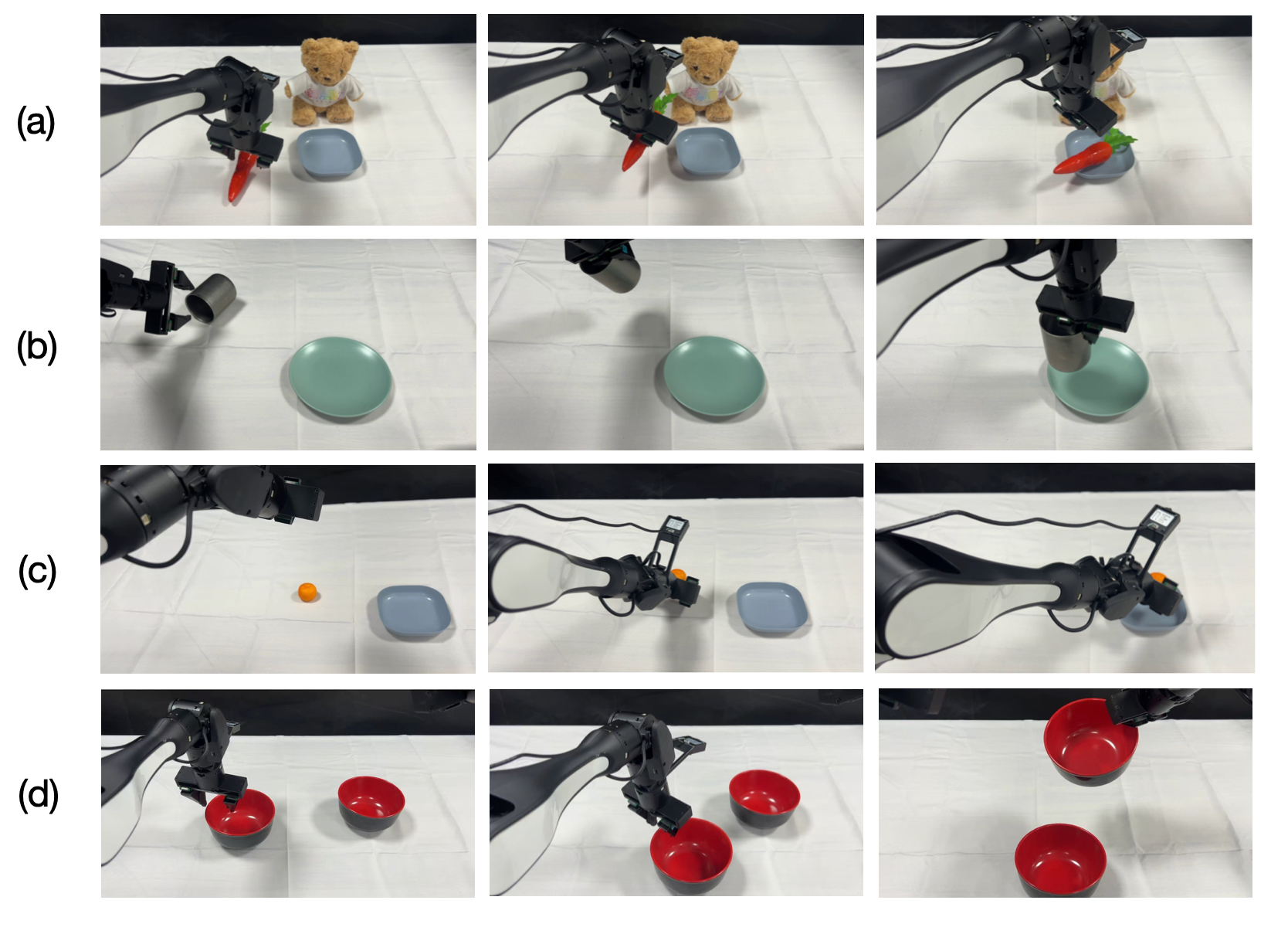} 
    \caption{
    \textbf{ALOHA robot experimental task demonstration. }Four representative tasks test MaskVLA's multi-view information integration capability: (a) Feed carrot; (b) Adjust cups; (c) Place orange; (d) Stack bowls.
    }
    \label{fig:real-setup}
\end{figure}

\subsection{Task Design}
We evaluate MaskVLA on four representative real-world manipulation tasks (Fig.~\ref{fig:real-setup}) to test multi-view perception and precise control:

\textbf{Feed carrot:} Place a carrot onto the food plate in front of a small bear. The elongated shape requires strong reliance on wrist-camera views during grasping and placement.

\textbf{Adjust cups:} Pick up cups on the desktop, align them upright, and place them at the center of a plate. Occlusions from the robotic arm necessitate effective use of wrist-camera observations.

\textbf{Place orange:} Pick up a small orange and place it in a fruit dish, evaluating precision and dexterity for small objects.

\textbf{Stack bowls:} Stack bowls using both arms, testing spatial understanding and bimanual coordination.

\subsection{Experimental Results Analysis}

\begin{figure}[t] 
    \centering
    \includegraphics[width=\linewidth]{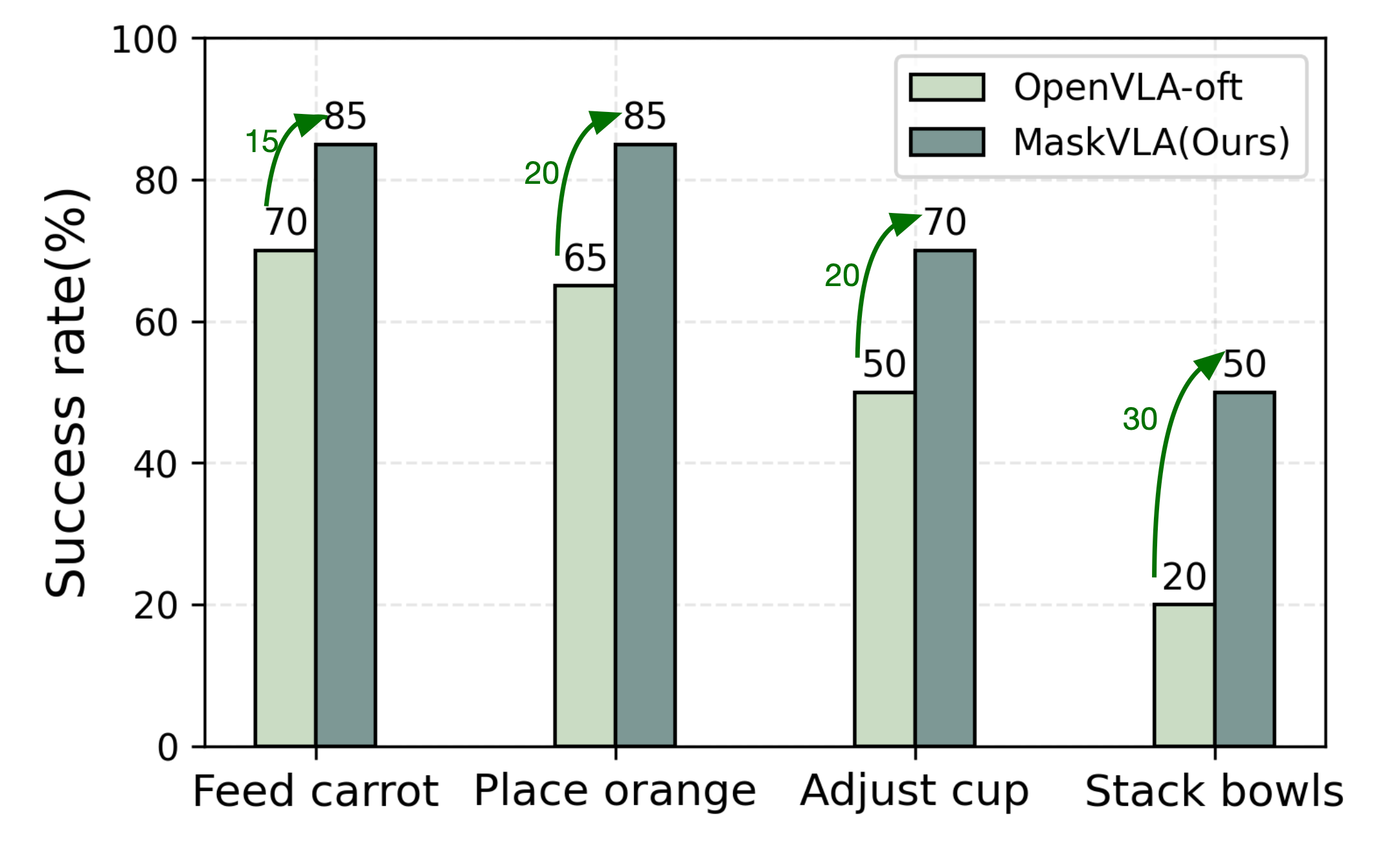} 
    \caption{
    \textbf{Real-world multi-task evaluation comparing the baseline and our method. }
    The bar chart shows task-wise success rates, highlighting the effectiveness of our approach in improving performance across all tasks under real-world conditions.
    }
    \label{fig:real-result}
\end{figure}

As shown in Fig.~\ref{fig:real-result}, MaskVLA consistently outperforms the state-of-the-art VLA fine-tuning method OpenVLA-OFT across all four real-world tasks. In particular, MaskVLA achieves success rates of 85.0\% on both the Feed carrot and Place orange tasks and 70.0\% in Adjust cup, outperforming the baseline. While OpenVLA-OFT demonstrates limited effectiveness in executing challenging, long-horizon manipulation tasks (e.g. Stack bowls), our MaskVLA shows substantial performance gains through balanced vision representation learning. Benefiting from the random masking strategy, MaskVLA effectively utilizes both the information from the primary view and wrist view to complete complex tasks and matigate the overfit problem, highlighting the enhanced visual generalization capability of our approach.

\section{CONCLUSIONS}

In this paper, we identify and address the overfitting issue in current Vision-Language-Action models (VLAs), which often compromises precise manipulation capabilities. To mitigate this, we propose \textbf{MaskVLA}—a novel fine-tuning method that incorporates a random masking strategy applied to the primary view, facilitating more balanced multi-view visual representation learning. Our approach effectively promotes active visual attention allocation and significantly improves manipulation performance. Extensive experiments in both simulated and real-world environments demonstrate the superiority of the proposed method. Furthermore, by evaluating various masking techniques, we find that applying minimal masking to the main camera view yields the best results.




\section*{Implementation Details}

Models were fine-tuned on 2 NVIDIA A100 GPUs (batch size 4 per GPU) for 30,000 steps per task ($\sim$12 hours). Experiments were conducted on RoboTwin2.0 with an RTX 3090 for evaluation and an A100 GPU for inference.

\section*{ACKNOWLEDGMENT}

This work was supported in part by the Guangdong S\&T Programme (Grant No. 2024B0101030002), the Basic Research Project of Hetao Shenzhen-HK ST Cooperation Zone (No. HZQBKCZYZ-2021067), the Shenzhen Outstanding Talents Training Fund (No. 202002), and the NSFC (Grant Nos. 62293482, 62573371, 61931024, and 12326610). 
This work was also supported by the Guangdong Province Radio Science Data Center (Grant No. 2025B1212070001), the Shenzhen General Program (No. JCYJ20220530143600001), the Guangdong Research Projects (Nos. 2017ZT07X152 and 2019CX01X104), the Guangdong Provincial Key Laboratory of Future Networks of Intelligence (Grant No. 2022B1212010001), the Shenzhen Key Laboratory of Big Data and Artificial Intelligence (Grant No. SYSPG20241211173853027), the National Key Research and Development Program of China (Nos. 2025YFF0515300 and 2025YFF0515304), the Open Project Program of Key Laboratory of Tibetan Information Processing, Ministry of Education (Grant No. QHSFCS-2606), the Shenzhen-Hong Kong Joint Funding (No. SGDX20211123112401002), and the Tencent \& Huawei Open Fund (Nos. 2024E0009 and 202301030019).




\printbibliography

\end{document}